# A NOVEL HISTOGRAM BASED ROBUST IMAGE REGISTRATION TECHNIQUE

V. Karthikeyan

Department of ECE, SVS College of Engineering
Coimbatore, India,
Karthick77keyan@gmail.com

*Abstract*-**In this paper, a method for Automatic Image Registration (AIR) through histogram is proposed. Automatic image registration is one of the crucial steps in the analysis of remotely sensed data. A new acquired image must be transformed, using image registration techniques, to match the orientation and scale of previous related images. This new approach combines several segmentations of the pair of images to be registered. A relaxation parameter on the histogram modes delineation is introduced. It is followed by characterization of the extracted objects through the objects area, axis ratio, and perimeter and fractal dimension. The matched objects are used for rotation and translation estimation. It allows for the registration of pairs of images with differences in rotation and translation. This method contributes to subpixel accuracy.**

**Keywords- Automatic Image Registration (AIR), Registration Techniques, Rotation and Translation Estimation**

## I. INTRODUCTION

Image registration refers to the task of aligning two or more images acquired at different times, from different sensors or from different viewpoints. Image registration is widely used in remote sensing, medical imaging, computer vision etc. In general, its applications can be divided into four main groups according to the manner of the image acquisition: Different viewpoints (multitier analysis) - Images of the same scene are acquired from different viewpoints and its main aim is to gain a larger 2D representation or a 3D representation. Different times (multitemporal analysis) **-** Images of the same scene are acquired at different times, often on regular basis, and possibly under different conditions and its main aim is to find and evaluate changes in the scene which appeared between the consecutive image acquisitions. Different sensors (multimodal analysis) **-** Images of the same scene are acquired by different sensors. The aim is to integrate the information obtained from different source streams to gain more complex and detailed scene representation. Image registration can roughly be classified into categories, namely, feature- and intensity-based techniques. Feature-based techniques depend upon detecting and matching landmark features within the images, and on the other hand, in intensity-based techniques, images are registered based on a relation between pixel intensity values of image. Automatic image registration is to perform the image registration task without the guidance and intervention of users. The large amount of incoming satellite images from the Earth Observing System (EOS) program and from new missions with hyper spectral instruments authorize the need for automatic image registration. One of the well-known techniques for image registration is to use Ground Control Points (GCPs). Most image segmentation methods can be classified according to their nature: histogram thresholding, feature space clustering, region-based approaches, edge detection approaches, fuzzy approaches, neural networks, physics-based approaches and any combination of these. Any of these generally intends to transform any image to a binary image: objects and background. Regarding histogram thresholding, several methods have been reported. Histogram-based image segmentation comprises three stages: recognizing the modes of the histogram, finding the valleys between the identified modes and finally apply thresholds to the image based upon the valleys. In this paper, a method for automatic image registration through histogram is proposed, which allows for a more detailed histogram-based segmentation than the other methods, and consequently to an accurate image registration. It estimates the rotation and translation between two images.

## II. METHOD DESCRIPTIONS

*A. Preprocessing*

To avoid undesirable segmentation results due to too much detail on the pixel domain it is advisable an image enhancement step before further processing. By image enhancement, it is intended to obtain an image with less detail than the original version. Image segmentation is to remove the degradation, which is assumed to be additive random noise. Histogram equalization of image 2 using the histogram counts of original image is performed, prior to the application of the Wiener filter. In this way, reduction of the image detail, as well as to the smoothing of the histogram is resulted

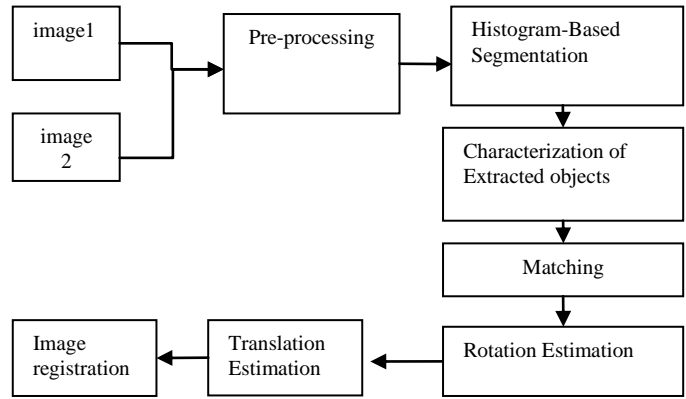

Fig.1. Block diagram of proposed AIR. through histogram

.

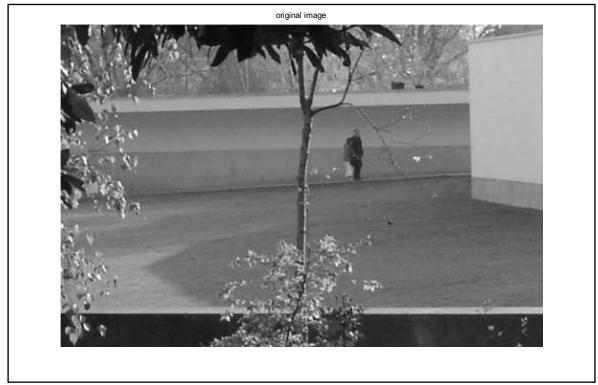

Fig.2. Image 1 (original image)

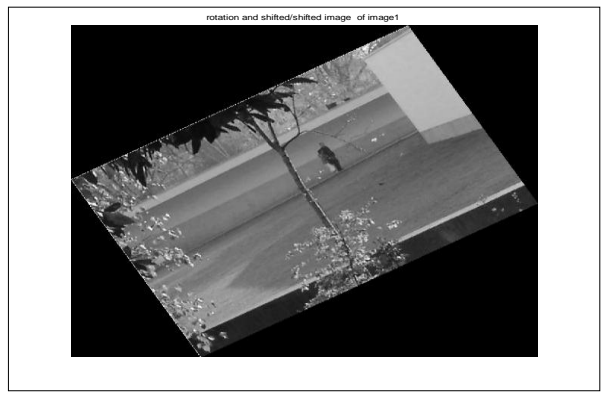

Fig.3 Image 2 (rotated and/or shifted image 1)

*B. Image Segmentation using histogram:*

The analysis of the consecutive slopes of the histogram is done. A relaxation parameter (α) is considered on the mode delineation, which in theory is a continuous parameter. It has to pass through a discretization process in practice. The inclusion of this parameter makes the subsequent stages of the proposed methodology to be more robust. The relaxation parameter (α) corresponds to the proportion of the height of the histogram considered to correspond to the highest mode. Below this value, the mode is to be considered as a "flat" region.

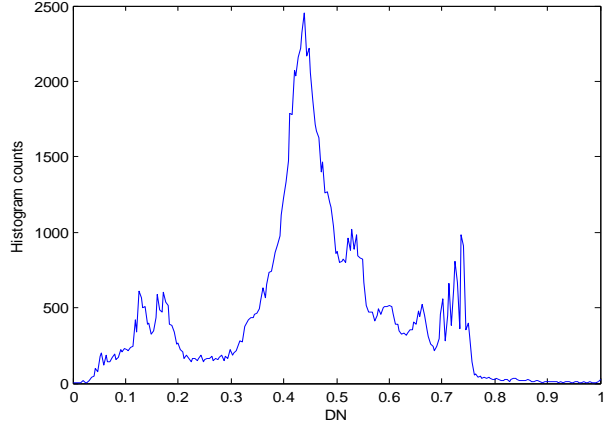

Fig..4 Histogram of the image presented in Fig.2 representing seven identified modes

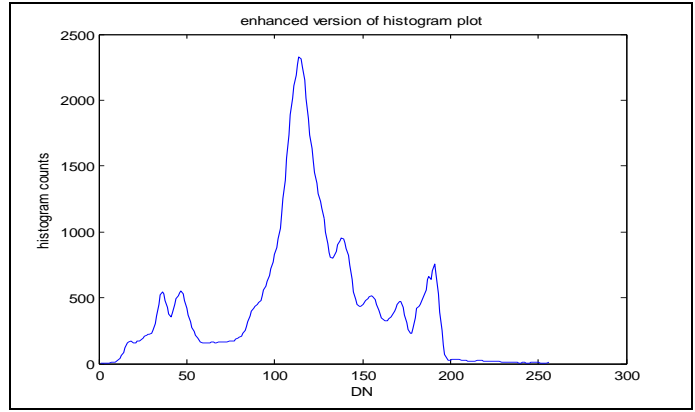

Fig. 5 Enhanced version of histogram plot of fig.4

*C. Extracted Objects and its characterization :*

The extracted objects are characterized by four attributes . They allow for their adequate morphological description: area($A_{rea}$) , perimeter($P_{erim}$) , axis ratio($A_{rt}$) and fractal dimension($D_b$).

*D. Matching*

The matching step is done with the evaluation of a cost function, between every possible two-by-two combination of objects obtained by the segmentation of the two images. The cost function, is defined as follows:

$$\gamma = \frac{(A_{rea1} - A_{rea2})}{\overline{A_{rea}}} + \frac{(A_{rat1} - A_{rat2})}{\overline{A_{rat}}} + \frac{(P_{erim1} - P_{erim2})}{\overline{P_{erim}}} + \frac{(D_{b1} - D_{b2})}{\overline{D_b}}$$

where $\overline{A_{rea}}, \overline{A_{rat}}, \overline{P_{erim}}$ and $\overline{D_b}$ are the average of each property for images 1 and 2 values.

*E. Rotation Estimation: $\hat{\theta}$*

The rotation and translation are determined on a statistical basis. The histogram of the extracted objects orientation differences is represented. $\hat{\theta}$ is found, through considering the frequencies of the rotation candidates, and finding the rotation value which absolute frequency corresponds to the higher occupant. This procedure leads to a robust estimation of θ.

*F. Translation Estimation:* $\hat{\delta}_x, \hat{\delta}_y$

Once $\hat{\theta}$ is obtained, only the initial matching candidates which correspond to the obtained rotation are considered. Then, a similar procedure as that followed in the rotation estimation is considered for obtaining $\hat{\delta}x$ and $\hat{\delta}_y$.

### III.   RESULTS AND DISCUSSION

In the presented examples, a pair of images with completely different histograms is accurately registered.

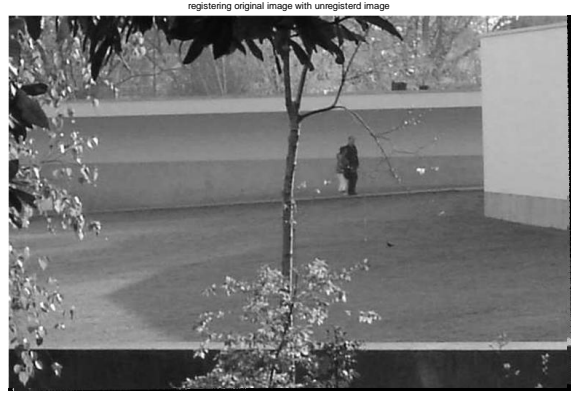
registering original image with unregistered image

Fig.6 Registration of original image with the simulated rotation and translation version of unregistered image.

TABLE I

| α (%) | Parameter | values for given image with reference values($30°$,60,40) |
|---|---|---|
| 20 | Rotation($\hat{\theta}$)<br>Translation in x ($\hat{\delta}_x$)<br>Translation in y ($\hat{\delta}_y$)<br>Elapsed time (sec) | $30°$<br>60<br>41.04<br>151 |
| 10 | Rotation<br>Translation in x<br>Translation in y<br>Elapsed time(sec) | $30°$<br>60<br>41.04<br>77.51 |
| 5 | Rotation<br>Translation in x<br>Translation in y<br>Elapsed time(sec) | $30°$<br>60<br>41.04<br>78.55 |

The proposed method is based upon detecting closed similar regions in both images and it will usually be possible to detect it, even for regions with low contrast. Furthermore, one important characteristic is the segmentation which is produced at different levels, by considering a range of values for the relaxation parameter. This allows for the robust subsequent stage of initial matching. Robust determinations of the rotation and translation parameters are determined on a statistical basis.  For larger images than those included in this work, the division of the images into tiles may be appropriate, since too much differences on the image content may difficult the application of the proposed methodology. The limit to consider the division of the image into tiles may depend upon several factors, such as the image content, among others. Moreover, the division of larger scenes into tiles, and considering the centers of each tile as matching points, it may become possible to correct for stronger distortions than the rigid-body transformation.

## IV. CONCLUSION AND FUTURE WORK

It was observed that the proposed method generally outperformed SIFT and the contour-based approach, in particular for the remote sensing examples. The implementation code of segmentation stage can be optimized in the future in order to provide a faster performance. The method was applied to single-band images at a time. However, in the future, adequate transformations of multispectral images to single-band images will certainly lead to even better results. The proposed methodology of image registration allowed for accurate results, even in the presence of noise. This method has shown to correctly register a pair of images at the subpixel accuracy.